\documentclass[french]{pfia} %,dochal
\usepackage[T1]{fontenc}
\usepackage[utf8]{inputenc}
\usepackage{comment}
\usepackage{subcaption}
\usepackage{enumitem}
\usepackage{graphicx}
\usepackage[dvipsnames]{xcolor}
\usepackage{url}
\definecolor{afiablue}{RGB}{61,159,207}
\definecolor{afiared}{RGB}{167,75,68}
\definecolor{afialightblue}{RGB}{158,193,232}

\newcommand{\field}[1]{\textcolor{afiablue}{#1}}
\newcommand{\val}[1]{\textsf{\textcolor{afialightblue}{#1}}}

\def\etc.{etc.\spacefactor=\the\sfcode`\c}
\newcommand{\cffig}[1]{(cf. figure~\ref{#1})}

\newcommand{\couleur}[2]{{\color{#1}#2}}

\newcommand{\sic}[1]{\og{}{\it #1 }\fg{}}

\newcommand{\iso}{{\sc ISO~30401}}
\newcommand{\isoQ}{{\sc ISO~9001}}

\newcommand{\guilli}[1]{\og{}{\it #1}\fg{}}
\newcommand{\lepdca}{{\couleur{PineGreen}{{\bf PDCA}}}}
\newcommand{\leseci}{{\couleur{RoyalBlue}{{\bf SECI}}}}
\newcommand{\pdca}[2]{{\couleur{PineGreen}{{\bf #1}{\it #2}}}}
\newcommand{\seci}[2]{{\couleur{RoyalBlue}{{\bf #1}{\it #2}}}}

\newcommand{\ittguilli}[2]{\item[$\triangleright$]\guilli{{\couleur{RoyalBlue}{#1}}}{ : #2}}
\newcommand{\ittguillj}[2]{\item[$\triangleright$]{\couleur{RoyalBlue}{#1} }{#2}}

\newcommand{\ProcPilotage}[1]{{\couleur{Red} {#1}}}
\newcommand{\ProcRealisation}[1]{{\couleur{Blue} {#1}}}
\newcommand{\ProcSupport}[1]{{\couleur{Green} {#1}}}

\newcommand{\guil}[1]{%
 \ifthenelse{\isempty{#1}}%
    {}% if #1 is empty
    {\og{}#1\fg{}}% if #1 is not empty
                    }

\title{\textbf{Intégration d'un système de management des connaissances conforme à l’ISO 30401 aux processus opérationnels existants d'une organisation}}

\author{Aline Belloni\fup{1} \& Patrick Prieur\fup{1}\\[6pt]
\fup{1} Ardans SAS,\\ 6 rue Jean Pierre Timbaud, \sic{Le Campus} Bâtiment B1, 78180 Montigny-le-Bretonneux, France\\
 \{abelloni, pprieur\}@ardans.fr $\bullet$        \url{https://www.ardans.fr}\\
 }
\date{12 juin 2025}

\begin{document}

\maketitle
\begin{resume}
Avec l'évolution des approches processus au sein des organisations, l'importance croissante des systèmes de management de la qualité (comme l'ISO 9001) et l'introduction récente de l'ISO 30401 pour le management de la connaissance, nous examinons comment ces différents éléments convergent dans la perspective d'un Système de Management Intégré. L'article démontre notamment comment un système de management des connaissances ISO 30401 peut être mis en œuvre en déployant les mécanismes du modèle \leseci ~au travers des étapes du cycle \lepdca ~tel qu'appliqué dans les processus du système de management intégré.
\end{resume}

\begin{motscles}
Système de Management des connaissances (SKM), \iso , SECI, PDCA, Modélisation des processus, Système de Management Intégré (SMI), métier.
\end{motscles}

\begin{abstract}
With the evolution of process approaches within organizations, the increasing importance of quality management systems (like ISO 9001), and the recent introduction of ISO 30401 for knowledge management, we examine how these different elements converge within the framework of an Integrated Management System. The article specifically demonstrates how a knowledge management system can be implemented by deploying the mechanisms of the \leseci ~model through the steps of the \lepdca ~cycle as applied in the processes of the integrated management system.
\end{abstract}

\begin{keywords}
Knowledge Management System, \iso , SECI, PDCA, Business process modelling, Integrated Management System (IMS).
\end{keywords}

% DEBUT DE L'ARTICLE

%=======================================================
\section{Introduction}
L’{\sc ISO~9000}:2015 \cite{iso9000} définit un processus comme un \sic{ensemble d'activités corrélées ou en interaction qui utilise des éléments d'entrée pour produire un résultat escompté}. L’approche processus dans une organisation consiste alors à identifier, décrire et regrouper sous forme de processus les activités et tâches qui concourent à l’atteinte des résultats de l’organisation, pour combiner ces processus en un Système de Management Intégré (SMI) reliant les niveaux opérationnels, pilotage et support selon la vision systémique développée par Le Moigne\cite{LeMoigne78}  ou de Rosnay\cite{deRosnay75} et reprise par la norme ISO~9000:2008\cite{iso2008}. Quelles soient certifiées ou simplement « alignées » avec cette norme, de très nombreuses organisations utilisent l’approche processus comme cadre essentiel pour garantir la « qualité » de leur fonctionnement, c’est-à-dire l'alignement des tâches réalisées avec leurs objectifs stratégiques (effectivité) et l'efficience du travail et des flux de travail de leurs collaborateurs.

Introduit en 2015 par l’ISO~9001 comme une exigence nouvelle du système de management de la qualité, le « sous-système » de management des connaissances au sein d’une organisation a été précisé en 2018 par la publication de la norme \iso ~dédiée \cite{iso30401} qui s’inscrit dans la famille des Normes de Systèmes de Management. Ces normes sont élaborées pour proposer un cadre normatif harmonisé aux processus les plus souvent rencontrés dans les organisations\footnote{La liste complète des Normes de Systèmes de Management, sectorielles ou non, est disponible sur le site de l’ISO \cite{isoMgt}. L’ISO a par ailleurs publié un manuel pour aider les organisations à intégrer les exigences de plusieurs NSM à leur propre système de management \cite{iumss2018}.}. En sa version actuelle (sa révision étant en cours), l’ISO 30401 prescrit quatre activités dédiées au développement des connaissances (cf. chapitre 4.4.2), à savoir, l’acquisition, l’application, la rétention et la gestion des connaissances obsolètes ou non conformes, et quatre activités et attitudes liées à la transmission et la transformation des connaissances (cf. chapitre 4.4.3) : interaction humaine, représentation, agrégation, assimilation et apprentissage. On constate que ces dernières sont directement inspirées du modèle \leseci ~de Nonaka et Takeushi \cite{Nonaka95} et \cite{Nonaka19} sur lequel nous revenons plus loin.

En tant qu’acteurs experts de la mise en œuvre de la norme \iso  nous sommes régulièrement confrontés à la nécessité d'expliquer aux équipes que nous accompagnons comment les activités de développement, de transmission et de transformation des connaissances décrites dans cette norme, s'intègrent aux processus opérationnels existants qui constituent le point de référence de ces équipes. A travers ces exercices, nous avons acquis la conviction qu’une organisation qui pratique réellement le cycle d’amélioration continue, prôné par l’\isoQ:2015, est en position favorable pour mettre en œuvre aisément les mécanismes du \leseci .

Basé sur notre expérience de mise en œuvre de l’\iso au cours de ces six dernières années dans des EPIC et grandes entreprises du secteur privé, cet article, après une rapide exploration des principes de modélisation des processus apportés par la norme \isoQ ~et du modèle \leseci , propose l’état de notre réflexion sur la manière d’articuler un système de management des connaissances (SKM) conforme à la norme \iso ~avec les autres processus d'un système de management intégré existant au sein de toute organisation.

\section{Cadre conceptuel pour la modélisation des processus métier}
L’approche processus fournit une cartographie des processus qui offre une vision globale des flux entre processus, ainsi qu’une représentation détaillée des activités et tâches relevant de chaque processus. 
\begin{figure}[ht]
    \centering
    \includegraphics[width=\textwidth/2]{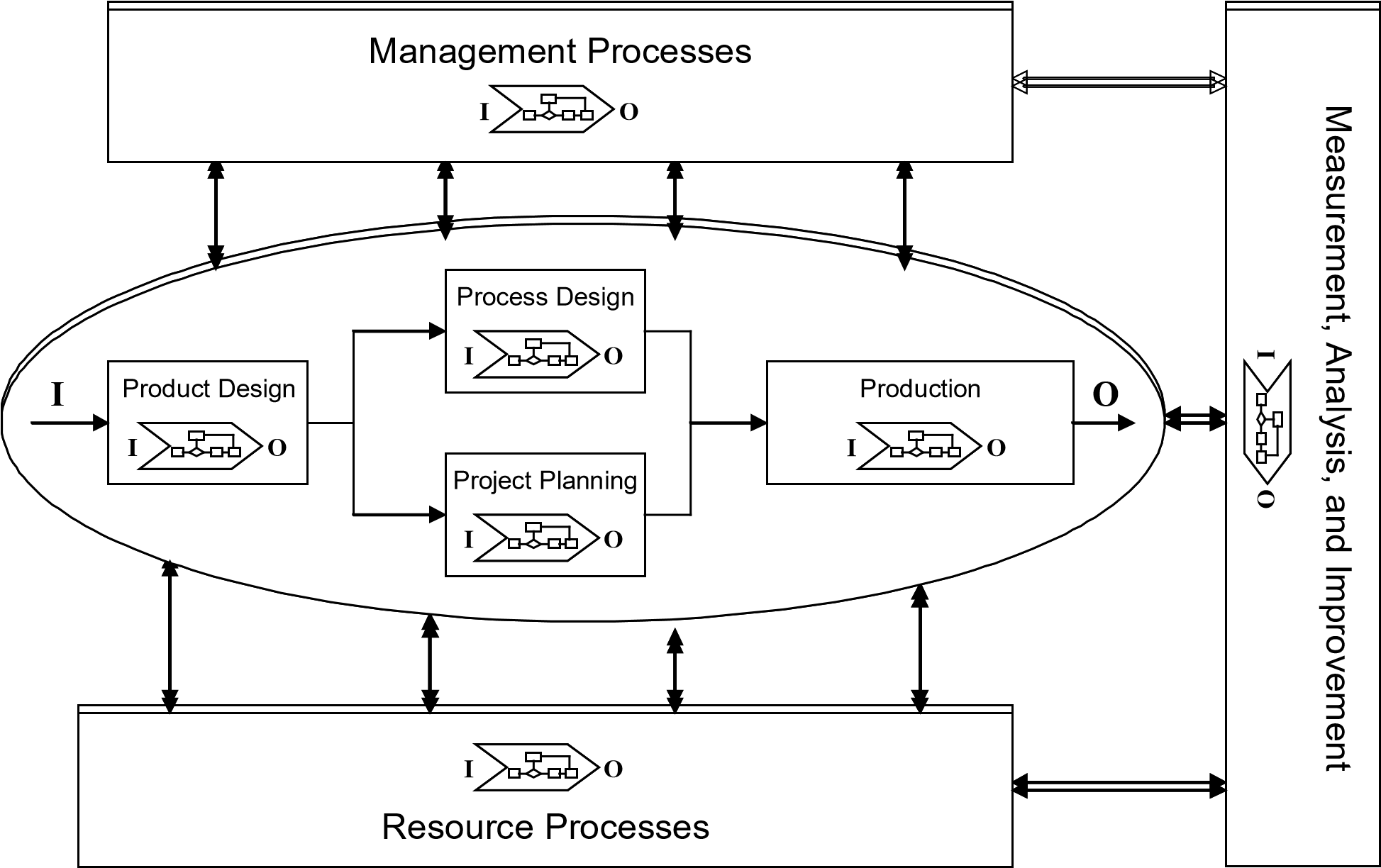}
    \caption{Une séquence de processus et ses interactions, selon \cite{iso2008}}
    \label{Fig_ArtApia_01}
\end{figure}
Le guide d’application de l’approche processus \cite{iso2008} paru conjointement à la famille de normes ISO 9000:2008 proposait de réaliser cette cartographie, illustrée ci-dessus \cffig{Fig_ArtApia_01}, en se basant sur une catégorisation en familles de processus : 
\begin{itemize}
    \item Processus de management (pilotage ou de {\it leadership})
    \item Processus de réalisation (processus opérationnels ou processus cœur de métier). Ces processus forment la chaîne de valeur de l’organisation
    \item Processus support (de soutien ou de ressources)
    \item Processus de mesure, d'analyse et d'amélioration. 
\end{itemize}
A noter que les processus de mesure sont souvent documentés comme des activités figurant directement à l’intérieur des processus de management, réalisation et support qu’ils concernent, tandis que les processus d’analyse et d’amélioration sont fréquemment traités comme des processus indépendants qui interagissent avec les autres processus, en admettant en entrée les résultats des activités de mesure et en générant comme produit de sortie, les instructions d’amélioration de ces autres processus.    

La figure ci-dessous \cffig{Fig_ArtApia_02} correspond à un exemple d’implémentation de cette approche observée dans un EPIC :

\begin{figure}[ht]
    \centering
    \includegraphics[width=\textwidth/2]{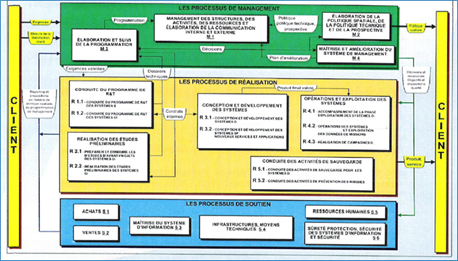}
    \caption{Une cartographie globale des processus selon \isoQ ~formalisée au sein d'un EPIC}
    \label{Fig_ArtApia_02}
\end{figure}

%~\cffig{Fig_ArtApia_03} (voir § \ref{Section_Exigences_User}).
%et ~\cfref{Tab_EGC25_User_SKM}.

Au cœur de la cartographie des processus se trouve le modèle « Activité » largement inspiré de méthodologies de modélisation de systèmes bien connues comme SADT \cite{galinier1989} ou IDEF0 \cite{IDEF0}. La norme \isoQ:2015 propose la représentation générique d’une activité ci-dessous :

%=======================================================
\begin{figure}[ht]
    \centering
    \includegraphics[width=\textwidth/2]{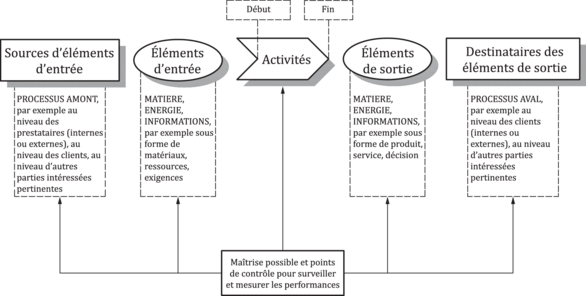}
    \caption{Représentation schématique des éléments d’un processus selon ISO~9001:2015 \cite{iso9001} : }
    \label{Fig_ArtApia_04}
\end{figure}

%Tiré de ISO 9001 :2015 [1] p viii « Figure 1 — Représentation schématique des éléments d’un processus »

Une ou plusieurs activités vont ainsi décrire un processus et chacune d’entre elles sera à son tour décrite par des tâches plus élémentaires. 

Par rapport à ce schéma, un pas de plus peut être fait en utilisant le célèbre « diagramme de la tortue » dont les composants sont introduits par Crosby dès 1979 \cite{Crosby1979}. 

Ce diagramme précise les éléments qu’il convient de détailler pour pleinement décrire un processus ou une activité : 
\begin{itemize}
 \item La carapace : Le processus lui-même pouvant être détaillé par les activités qui le composent
 \item La tête : Les entrées ou \sic{inputs} à transformer par le processus
\item La queue : Les sorties ou \sic{outputs} du processus (produits ou services)
\item Les pattes : Les ressources nécessaires (qui, avec quoi), les méthodes (comment) et les indicateurs de performance (combien)
\end{itemize}

\begin{figure}[ht]
    \centering
    \includegraphics[width=\textwidth/2]{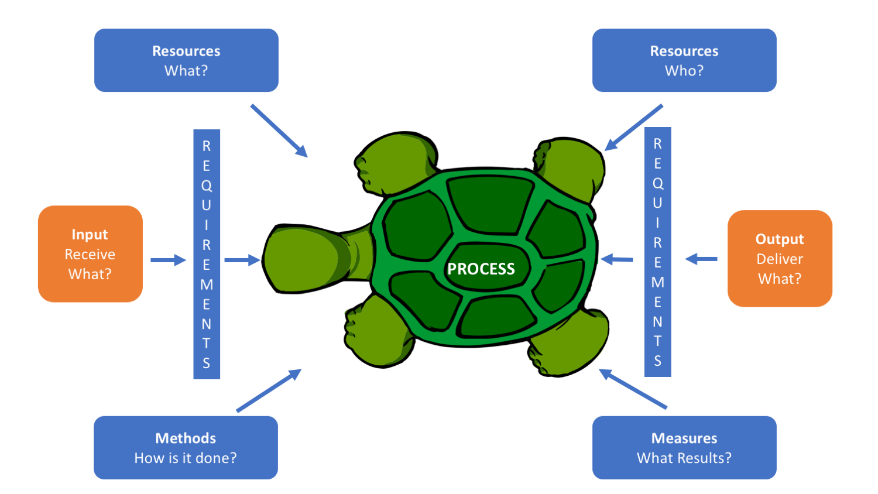}
    \caption{\sic{How to use Turtle Diagrams} selon IATF16949 Standards Store}
    \label{ArtApia_Turtle}
\end{figure}

La figure ci-dessous \cffig{ArtApia_Turtle} illustre le diagramme de la tortue\footnote{\url{https://16949store.com/articles/how-to-use-turtle-diagrams/}}.

L’intérêt de ce type de diagramme, du point de vue du management des connaissances, est qu’il cartographie {\it de facto} les connaissances explicites nécessaires à l’exécution du processus, sans nécessairement les détailler : savoirs relatifs aux entrées, sorties et contraintes d’environnement (le quoi ?) mais aussi savoir-faire à mettre en œuvre pour opérer la transformation des entrées en sorties (qui ? avec quoi ? comment ?).

\section{ISO9001:2015, Au-delà du simple “Faire” (Do)  : le cycle PDCA associé au pilotage et support du système de management}\label{sect:DoPDCA}

Le cycle \lepdca ~ou « roue de Deming » \cite{Deming} est le cycle d'amélioration continue adopté par l’ISO 9000 dès sa version 2000. Son principe \sic{{PLAN $\rightarrow$ DO  $\rightarrow$ CHECK $\rightarrow$ ACT}} consiste à planifier une activité (« Plan »), à faire (« Do ») c’est-à-dire mettre en œuvre l’activité planifiée, à évaluer (« Check ») les résultats obtenus et à prendre les mesures appropriées pour ajuster (« Act ») le cas échéant le cycle suivant en fonction des conclusions observées.

S'appuyant sur la version de 2008, la norme \isoQ:2015 encourage les organisations à adopter une approche holistique de la modélisation de leurs processus en soulignant que la description d'un processus ou d'une activité ne consiste pas simplement à décrire l’activité de réalisation (le « DO »), mais à décrire l’ensemble du système qui concourt à cette réalisation, ce qui implique de rajouter la description de la boucle d’amélioration continue du processus mais aussi les processus de pilotage ou \sic{leadership} et \sic{support} comme vu précédemment dans le cadre de la vision systémique de l’approche processus :

\begin{itemize}
\item Planifier-Réaliser-Vérifier-Agir (ou \sic{Ajuster}) 
\item Piloter
\item Soutenir 
\end{itemize}

Sur ce principe, l’\isoQ :2015 propose la représentation générique suivante de tout système de management de la qualité \cffig{Fig_ArtApia_03}: 
\begin{figure}[ht]
    \centering
    \includegraphics[width=\textwidth/2]{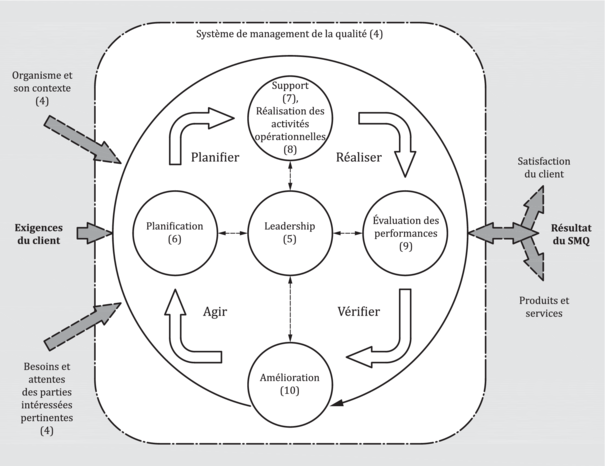}
    \caption{Représentation de la structure de tout Système de Management dans le cycle \lepdca ~selon \cite{iso9001} }
    \label{Fig_ArtApia_03}
\end{figure}

Suivant le principe fractal de la modélisation systémique, le cycle d’amélioration continue se retrouve à tous les niveaux d'un processus, de la vue d'ensemble la plus large à l’étape la plus granulaire (tâche). Notons cependant que le degré de formalisation de l'application du \lepdca ~varie en conséquence, de généralement très formalisé et détaillé au niveau global du processus\dots{} à souvent totalement informel au niveau le plus élémentaire de la tâche exécutée par un individu. 

\begin{figure}[ht]
    \centering
    \includegraphics[width=\textwidth/2]{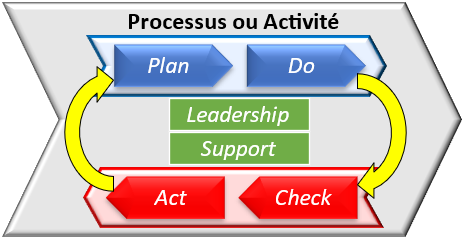}
    \caption{Reformulation de la représentation générique d'un processus}
    \label{ArtApiaProcAct}
\end{figure}
Si nous adoptons la représentation générique   \cffig{ArtApiaProcAct} d'un processus, d'un sous-processus ou d'une activité,
du point de vue de la cartographie globale des processus, remarquons que l’on peut continuer à regrouper tout ou partie des activités de «~Leadership~» et de «~Support~» dans des processus appartenant aux familles de « Pilotage » et « Support » de l’approche systémique évoquées ci-avant :
\begin{tabular}{p{0.5\textwidth}}%\setlength{\tabcolsep}{1pt}%\setlength{\arrayrulewidth}{0mm}
$\triangleright$ \ProcRealisation{L’activité « Leadership » d’un processus de réalisation}\\ devient \ProcPilotage{l’activité « Réalisation » d’un processus de Pilotage}.\\
$\triangleright$ \ProcRealisation{L’activité « Support » d’un processus de réalisation}\\devient \ProcSupport{l’activité « Réalisation » d’un processus de Support}.\\
\end{tabular}

Ce qui aboutit à la représentation générique ci-dessous d’un Système de Management Intégré que nous utiliserons plus loin pour intégrer les processus de management des connaissances.

\begin{figure}[ht]
    \centering
    \includegraphics[width=\textwidth/2]{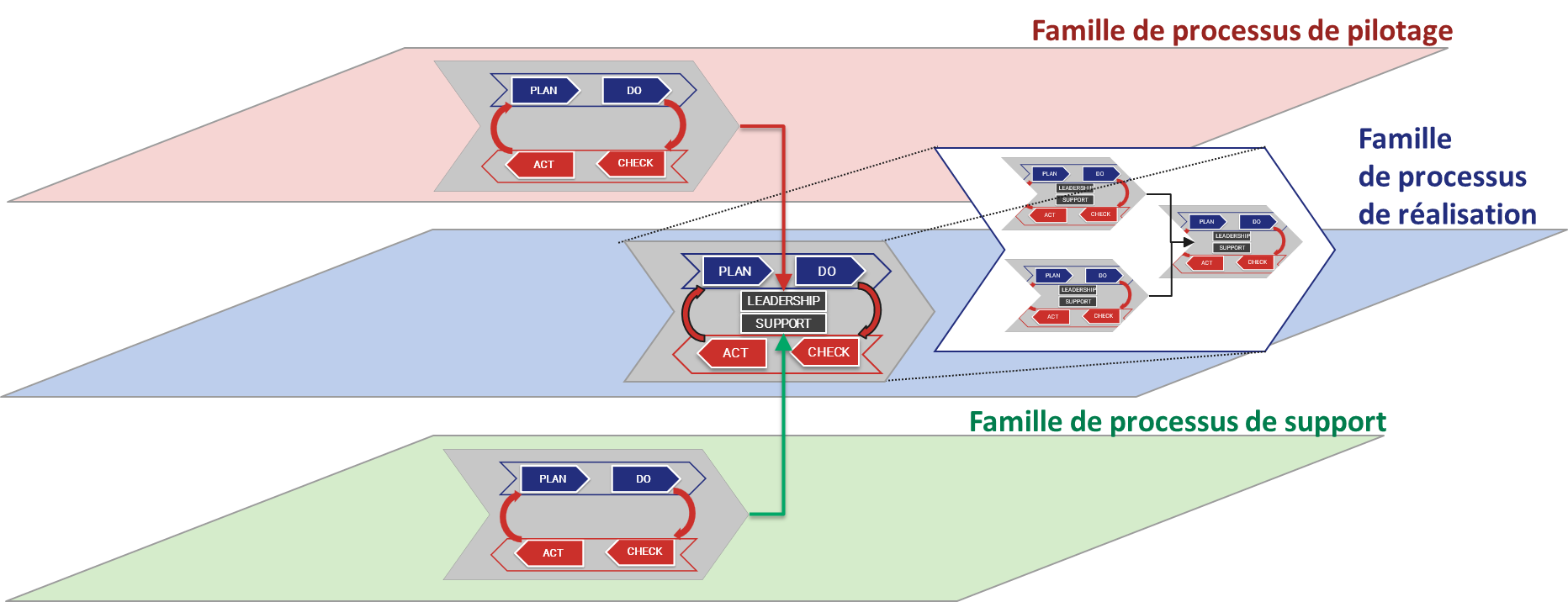}
    \caption{Cartographie générique des processus d’un système de management intégrant le \lepdca }
    \label{ArtApiaCartoGenerique}
\end{figure}

Notons que les processus de pilotage et de support d’un système de management sont des processus à part entière et possèdent donc eux-aussi des sous-processus de pilotage et de support que nous ne faisons pas figurer ici par souci de simplification \cffig{ArtApiaCartoGenerique}.

%%XXXXXXXXXXXXXXXXXXXXXXXXXXXX

\begin{figure}[ht]
    \centering
    \includegraphics[width=\textwidth/2]{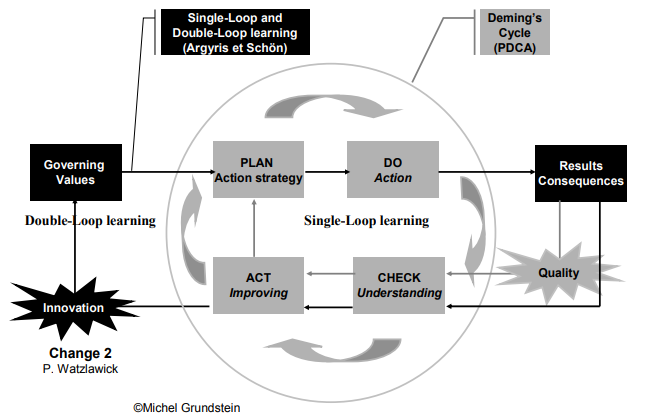}
    \caption{Deming’s cycle and Argyris \& Schön’s Organizational learning, tel que vu par Michel Grundstein \cite{Grundstein2012} }
    \label{ArtApia2LoopMG}
\end{figure}

Michel Grundstein \cite{Grundstein2012} \cffig{ArtApia2LoopMG} établit dès 2012 un pont entre le \lepdca ~comme moteur de l’amélioration continue de l’organisation, et le KM comme moteur de l’organisation apprenante, en observant que le \lepdca ~correspond précisément à la notion de simple  boucle d’apprentissage (\sic{Single loop learning}) développée par Argyris et Schön \cite{Argyris2002} tandis que la capacité à déployer la seconde boucle décrite par ces mêmes auteurs (\sic{Double-loop learning}) correspond, ou tout du moins s’intègre, à ce que l’on nomme couramment « innovation » dans les organisations. 
Sur cet aspect, Grundstein relève la convergence avec le concept de « Changement 2 » (changement de la loi de composition interne qui gouverne le système) développé par Watzlawick et al. \cite{Watzlawick1975}.
.

Notre observation est que le \lepdca, lorsqu’il est correctement pratiqué, a effectivement tendance à assez bien assurer la simple boucle d’apprentissage en faisant émerger des ajustements correctifs et préventifs vis-à-vis de savoirs et savoir-faire établis ayant entraîné, par leur utilisation, des résultats non conformes ou non souhaités, ceci conduisant à la mise à jour des référentiels de connaissances métier. 

Par contre, même si bien entendu nous observons désormais systématiquement des processus d’innovation mis en place pour favoriser l’éclosion d’innovations incrémentales ou de rupture, nous constatons plus rarement que ces processus soient conçus pour se nourrir d’une capacité institutionnelle à s’élever systématiquement par rapport à la simple boucle d’apprentissage afin d’entrer dans cette seconde boucle qui vise à remettre en cause les valeurs et principes qui avaient conduit à l’établissement des savoirs et savoir-faire utilisés. C’est précisément ce manque que, selon nous, le processus de management des connaissances peut venir combler en déployant les mécanismes du modèle \leseci , complémentairement aux cycles \lepdca ~des processus de l’organisation. 

\section{Le modèle SECI ou cycle de la création des connaissances }
En 1995, Ikujirō Nonaka et Hirotaka Takeuchi \cite{Nonaka95} proposaient le modèle \leseci , cycle de création de la connaissance dans une organisation, composé de quatre mécanismes : la socialisation « \seci{\bf S}{} », l’externalisation « \seci{\bf E}{} », la combinaison « \seci{\bf C}{} » et l’internalisation (ou intériorisation) «~\seci{\bf I}{}~» de la connaissance. Les auteurs font évoluer leur modèle en 2019 \cite{Nonaka19} en une spirale de création ET d’application (\sic{practice}) de la connaissance en introduisant les dimensions temporelle et sociétale (\sic{ontological}) et en positionnant la {\it phronesis} aristotélicienne (\sic{practical wisdom}) présente en chaque individu comme force motrice de cette spirale. 
 
Ce modèle repose sur l’idée que les connaissances émergent, circulent et sont transformées, en étant appliquées au sein des organisations par les collaborateurs, à travers un processus de conversion de la connaissance tacite individuelle en connaissance explicite organisationnelle, partagée collectivement, et ainsi de suite.
\begin{figure}[ht]
    \centering
    \includegraphics[width=\textwidth/2]{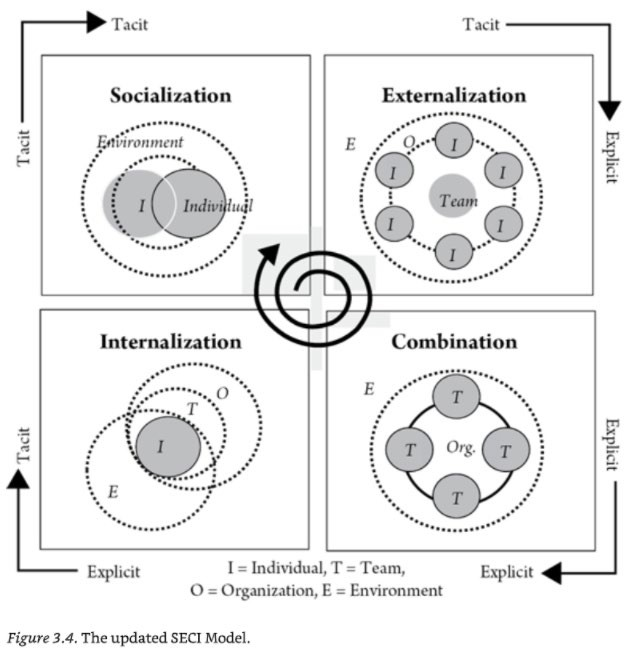}
    \caption{The updated \leseci ~Model, extrait de \cite{Nonaka19} }
    \label{ArtApiaSECI}
\end{figure}
%%XXXXXXXXXXXXXXXXXXXXXXXXXXXX
Le schéma \cffig{ArtApiaSECI} issu de \cite{Nonaka19}, fait apparaître les quatre mécanismes du \leseci ~correspondant à des combinatoires spécifiques des formes tacite ou explicite que prend la connaissance et des vecteurs de cette connaissance allant de l’individuel au collectif (individu, groupe d’individus, organisation).

La socialisation « \seci{\bf S}{} » et l’internalisation « \seci{\bf I}{} » touchent plus particulièrement aux connaissances tacites que les individus acquièrent  par l’expérience personnelle vécue ou l’apprentissage et qu’ils enrichissent par les échanges entre collègues, tandis que l’externalisation « \seci{\bf E}{} » et la combinaison « \seci{\bf C}{} » concernent la conversion des connaissances tacites en explicites (ou codifiées) qui sont produites et validées par le collectif (groupes d’individus ou organisation) pour être valorisées à travers l’usage que les individus en font dans leurs situations opérationnelles après les avoir internalisées (rendues tacites).

Si la littérature recense nombre d’exemples d’implantation du modèle \leseci ~comme par exemple dans \cite{Nishihara-2018CD} et \cite{Nishihara-2018PA}, nous n’avons pour notre part, dans nos diagnostics préalables à la mise en œuvre du management des connaissances, jamais véritablement rencontré en France d’organisation ayant institutionnalisé l’application de l’ensemble de ce modèle. 
Il est, par contre, courant d’en constater un déploiement partiel notamment relativement aux mécanismes de socialisation et d’internalisation :  ainsi le passage de relais ou le biseau mis en place lors d’un départ annoncé, l’existence de communautés de pratique dédiées à des thématiques ou sujets métier importants pour l’organisation, l’existence d’annuaires d’expertise visant à trouver et solliciter le bon expert lors de la survenue d’un problème ou encore l’accompagnement ou le coaching d’un nouveau collaborateur lors de son \sic{on-boarding} sont des exemples rencontrés.

\begin{figure}[ht]
    \centering
    \includegraphics[width=\textwidth/2]{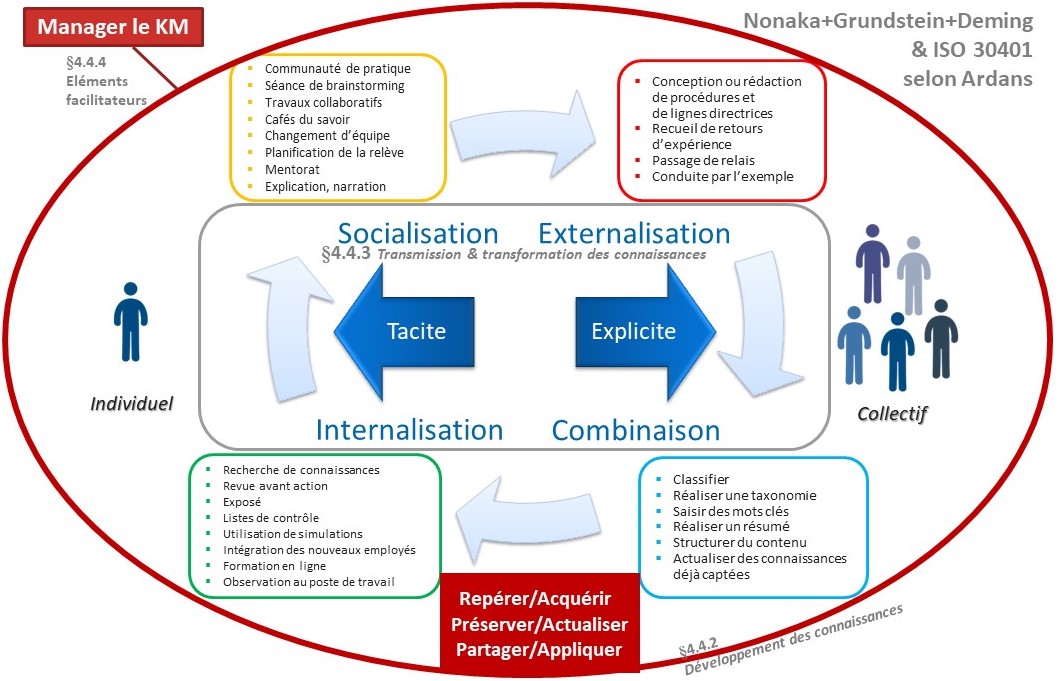}
    \caption{La consistance de l'ISO30401 face aux principes de Nonaka et Grundstein en particulier~\cite{ArdansIC2023}} %ArdansIC2023
    \label{Fig_ISO}
\end{figure}

A partir de 2015, l’\isoQ ~exige, dans son paragraphe 7.1.6, de démontrer la gestion des « connaissances organisationnelles » définies comme les « connaissances nécessaires à la mise en œuvre de ses processus et à l’obtention de la conformité des produits et des services ». Nombreuses sont alors les organisations qui, ne sachant pas comment répondre à cette exigence, sont sanctionnées par une non-conformité dans le cadre d’un audit de re-certification. Le modèle \leseci ~repris au chapitre 4.4.3 de l’\iso, offre alors une nouvelle lecture du 7.1.6 en précisant quelles connaissances doivent être gérées : les tacites, les explicites, celles de l’individu mais aussi celles du collectif… tout en listant pour chaque étape du \leseci ~un certain nombre d’outils, méthodes ou comportements favorables à la mise en œuvre de ces mécanismes. Dans la figure \cffig{Fig_ISO}, nous repositionnons les « activités et comportements » listés par l’\iso ~par rapport au \leseci .
Néanmoins l’\iso ~reste flou quant à la traduction du \leseci, dont {\it in fine} on souhaite pouvoir \textit{constater} la présence et les effets dans l’organisation, en différents processus que l’on souhaite objectivement pouvoir \textit{déployer}, tels que listés dans le chapitre 4.4.2 de la norme. 

\section{Intégration du management des connaissances au sein d’un système de management intégré : le cycle PDCA et le modèle SECI}

À travers plusieurs exemples d’implantations de SKM dans un environnement compatible \isoQ , nous avons observé que le cycle \lepdca ~et le modèle \leseci ~pouvaient être naturellement rapprochés et combinés.

\subsection{Modèle SECI ~dans les étapes «\texorpdfstring{\pdca{P}{lanifier}}{Planifier} » (P) et «\texorpdfstring{\pdca{F}{aire}}{Faire} » (D) du cycle PDCA }

%\subsection{Modèle SECI ~dans les étapes «~\pdca{P}{lanifier}~» ({\bf P}) et «~\pdca{F}{aire}~» ({\bf D}) du cycle PDCA}

Ces deux premières étapes, intrinsèquement opérationnelles, exigent que les individus disposent des connaissances nécessaires pour les exécuter efficacement. Il est donc indispensable de les «~responsabiliser~» en favorisant les aspects «~\seci{\bf I}{}~» (\seci{I}{nternalisation}) et «~\seci{\bf S}{}~» (\seci{S}{ocialisation}) du modèle \leseci ~(que l'on retrouve assimilation et apprentissage et interaction humaine du 4.4.3a et b de l’\iso). Cela implique la mise en place et le maintien d'environnements technologiques et sociaux garantissant l'accessibilité, le partage et la diffusion des connaissances, qu’elles soient explicitées ou tacites, permettant ainsi aux individus de préparer et exécuter efficacement leur travail.
\begin{figure}[ht]
    \centering
    \includegraphics[width=\textwidth/2]{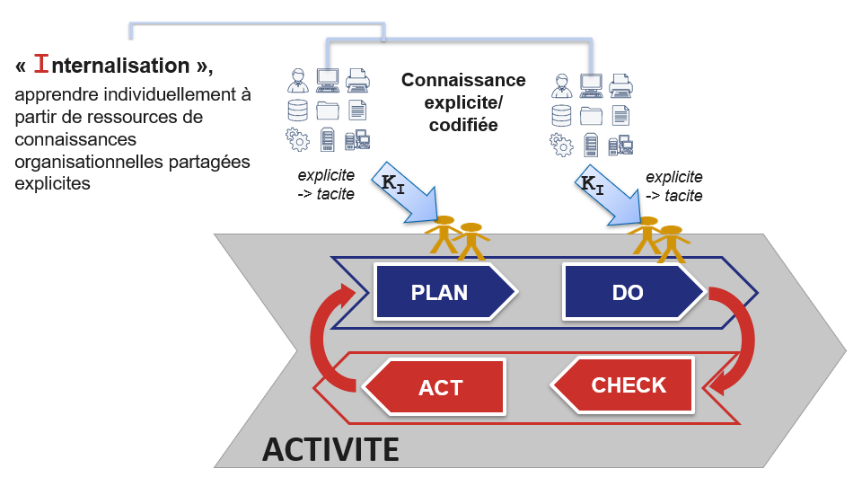}
    \caption{\seci{I}{nternalisation} et le \lepdca}
    \label{Fig_ArtApia_07}
\end{figure}
\begin{itemize}
\ittguilli{\seci{\bf I}{}nternalisation}{les individus apprennent à partir de connaissances organisationnelles explicites partagées au sein de documents (normes, procédures opérationnelles, règles, référentiels, standard…), mais aussi via des systèmes d'apprentissage en ligne ou encore à l’aide de formateurs reconnus, convertissant ces nouvelles connaissances explicites en connaissances tacites, grâce à leur expérience personnelle dans un contexte professionnel donné. Cette étape permet à l'organisation de vérifier si son capital de connaissances explicites collectives est effectivement disponible et appliqué par ses collaborateurs lors de la réalisation d’activités métier.}
\ittguilli{\seci{\bf S}{}ocialisation}{les individus apprennent, comparent et génèrent de nouvelles idées et connaissances grâce au partage de connaissances tacites, par le biais d'interactions directes, notamment au sein d'équipes transversales (compagnonnage par exemple), de communautés de pratique ou de réseaux métier. Cette étape permet également à l'organisation de s'assurer que ses actifs de connaissances tacites collectifs sont mis à disposition et exploités, même si le suivi de cet effet est plus complexe.}
\end{itemize}
\begin{figure}[ht]
    \centering
    \includegraphics[width=\textwidth/2]{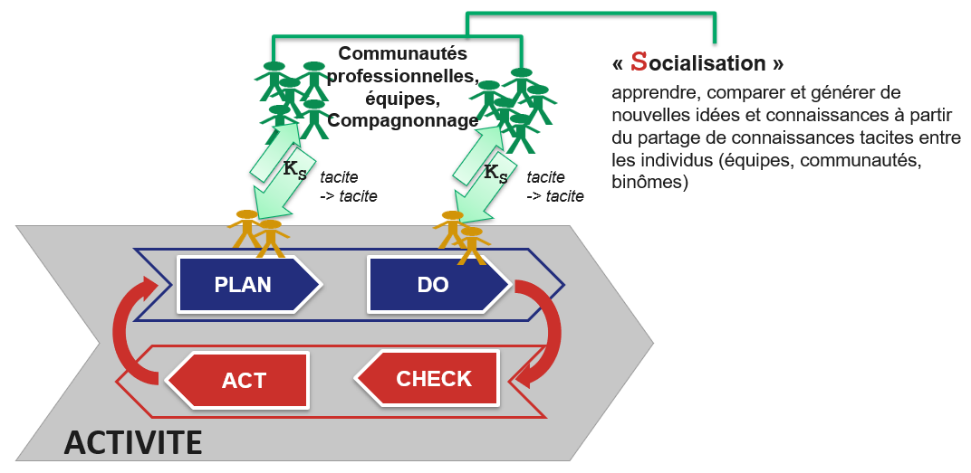}
    \caption{\seci{S}{ocialisation} et le \lepdca}
    \label{Fig_ArtApia_08}
\end{figure}

% \begin{tabular}{lp{\ImgTailleO}}
% $\triangleright$ &\itemtt{Agréger} dans des codes de calculs les fruits des expériences analysées et de leur modélisation associées,\\
% $\triangleright$ &\itemtt{Numériser} des documents (avant qu’il ne s’effacent), les référencer et les archiver selon les règles,\\
% $\triangleright$ &\itemtt{Filmer} les gestes métiers appropriés dans les opérations manuelles,\\
% $\triangleright$ &\itemtt{Recueillir {et}  expliciter} les retours d’expériences, les savoirs, les expertises des sachants avant qu’ils ne quittent leurs fonctions.\\
% \end{tabular}

\subsection{Modèle SECI ~dans les étapes «\texorpdfstring{\pdca{V}{érifier}}{Vérifier} » (C) et «\texorpdfstring{\pdca{A}{gir}}{Agir} » (A) du cycle PDCA }
%\subsection{Modèle SECI ~dans les étapes « \pdca{V}{érifier} » (C) et « \pdca{A}{gir} » (A) du cycle PDCA }
Contrairement aux deux premières, il s'agit ici d'étapes collectives et réflexives visant à remettre en question (confirmer ou invalider) et à améliorer les connaissances existantes qui ont été appliquées, ainsi que les connaissances nouvellement créées pour conduire à leur explicitation. 

Les activités «~\seci{\bf E}{}~» \seci{E}{xternalisation} et «~\seci{\bf C}{}~» \seci{C}{ombinaison} du \leseci ~garantissent que les étapes de vérification «~\pdca{\bf C}{}~» et d'ajustement «~\pdca{\bf A}{}~» du \lepdca ~sont réalisées de manière «~orientée connaissances~». Selon les résultats observés, les nouvelles connaissances créées seront explicitées, validées et combinées aux ressources de connaissances existantes, afin d’enrichir au fil de l’eau le patrimoine de connaissances de l’organisation.

\begin{figure}[ht]
    \centering
    \includegraphics[width=\textwidth/2]{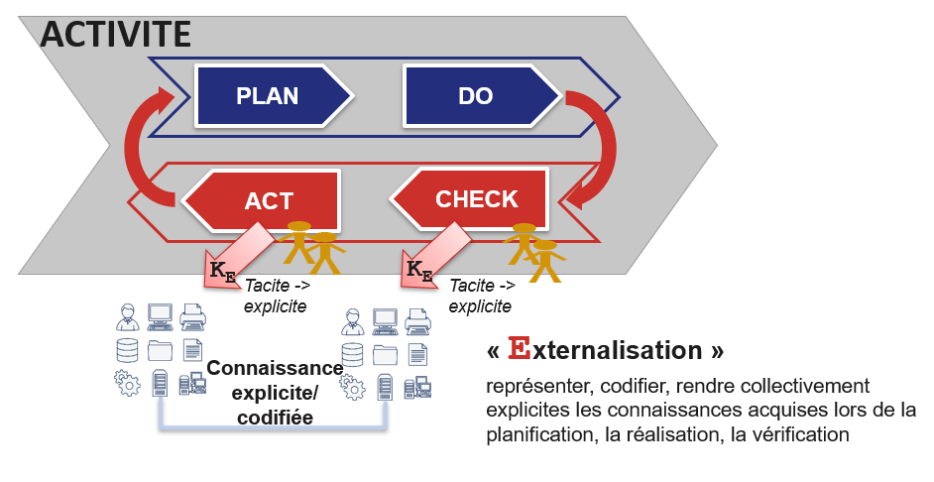}
    \caption{\seci{E}{xternalisation} et le \lepdca}
    \label{Fig_ArtApia_09}
\end{figure}

\begin{itemize}
\ittguilli{{\bf E}xternalisation}{les connaissances utilisées lors de la planification et de l'action, qu'elles soient issues de connaissances potentielles existantes ou créées {\it de facto}, sont révélées, complétées, explicitées dans leur contexte. En effet, avec l'étape «~Vérifier~» «~\pdca{\bf C}{}~», les individus réfléchissent collectivement à leurs propres expériences vécues et aux résultats observés des connaissances tacites ou explicites appliquées, afin de comprendre et de formaliser de nouvelles connaissances ou d’améliorer des connaissances existantes (étape «~\pdca{\bf A}{}~» «~Agir~» ), comme par exemple la compréhension consensuelle des causes profondes d'un échec ou d'une réussite.

Ce type d’exercice est communément pratiqué sous le vocable de «~retour d’expérience~» ou \sic{lessons identified / lessons learned}.}
\ittguilli{{\bf C}ombinaison}{les connaissances sont élevées au rang de connaissances applicables, collectivement organisées et validées. Le principe fractal du modèle (mêmes principes appliqués aux niveaux des tâches, des activités et des processus) garantit qu'au fil du temps, les connaissances passent du statut de leçons apprises sur le terrain ou de connaissances générées sur le terrain, à celui de connaissances validées par les métiers et de meilleures pratiques reconnues au plus haut niveau (\sic{ Corporate Knowledge of the Art}).}
\end{itemize}
\begin{figure}[ht]
    \centering
    \includegraphics[width=\textwidth/2]{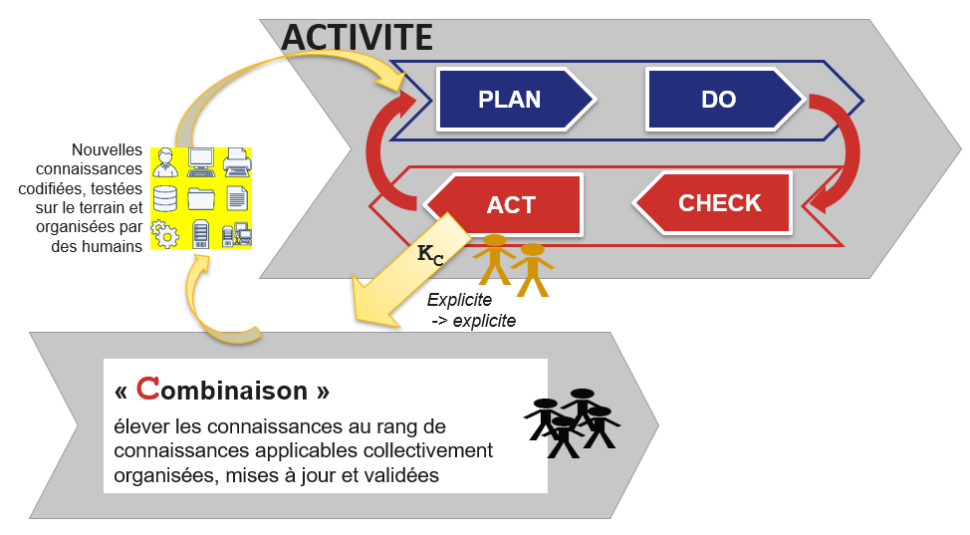}
    \caption{\seci{C}{ombinaison} et le \lepdca}
    \label{Fig_ArtApia_10}
\end{figure}
Fort de ces quatre rapprochements entre cycle \lepdca ~et modèle \leseci, nous traduisons concrètement cela dans une vision holistique du système de gestion des connaissances intégré au sein d’un Système de Management Intégré (SMI), dès lors que deux processus complémentaires et essentiels sont pris en compte : le \sic{leadership} et le \sic{support} aux activités de gestion des connaissances.
%===============================================================
\section{La vision holistique des processus du système de management des connaissances}
Pour être conforme à la vision systémique de l’approche processus, l'intégration par une organisation d'un système de management des connaissances (SKM) au sein de son SMI implique, outre ce que nous venons de décrire plus haut au niveau opérationnel, la prise en compte des deux sous-processus de pilotage et de support au management des connaissances.

%\begin{enumerate}
 %   \item 
 \paragraph{{\bf 1. «~Processus de pilotage du management des connaissances~» }}
    
    Ce processus, rattaché à la famille de processus de «~Pilotage~», se concentre sur l’élaboration du plan d’actions KM et le pilotage de la mise en œuvre de ce dernier. Ce processus comporte les activités suivantes :
    %\begin{enumerate}
        \begin{itemize}
        \ittguilli{L’établissement d’une politique KM}{ définition de l’approche de l’organisation en matière de création, de partage et d’utilisation des connaissances, en l’alignant sur les objectifs et la stratégie de l’organisation~;}
        \ittguilli{L’élaboration et la mise en œuvre d’un cadre de gouvernance des connaissances}{ désignation de l’organisation, rôles et responsabilités permettant de garantir une application cohérente des principes de management des connaissances à tous les niveaux et fonctions~;}
        \ittguilli{L’intégration des principes du modèle \leseci ~au cycle \lepdca}{reconnaissance du fait que les quatre étapes socialisation, externalisation, combinaison, internalisation, sont essentielles tout au long du cycle \lepdca , sans se limiter à certaines étapes spécifiques~;}
        \ittguilli{La définition d’un plan d’actions KM}{déclinaison de la politique, à partir notamment de l’élaboration d’une cartographie des connaissances clés et vulnérables pour identifier et prioriser les actions à mener~;}
        \ittguilli{La définition des actions de transformation culturelle et technique nécessaires}{gouvernance et planification des actions pour l’adoption des pratiques de management des connaissances au sein de l'organisme.}
        \end{itemize} %===============================================================
    %\end{enumerate}

%\item
\paragraph{{\bf 2. «~Processus de support au management des connaissances~»}} 

Ce processus, rattaché à la famille des processus « support » concerne notamment :
%\begin{enumerate}
    \begin{itemize}
    \ittguilli{La mise à disposition des méthodes et ressources KM (humaines et techniques) pour «~activer~» les mécanismes d'internalisation et de socialisation des connaissances}{par exemple les programmes de mentorat au sein de communautés de pratique, la recherche intelligente dans les référentiels de connaissances métier, etc. Ceci, pour faciliter la création, le partage et l'apprentissage des connaissances par les individus, l'intégration des connaissances existantes dans la prise de décision et le suivi de leur application à toutes les étapes \lepdca , et plus particulièrement le «~\pdca{\bf PLAN \&  DO}{}~» de tous les processus et activités métier ciblés.}
    \ittguilli{La mise à disposition des méthodes et ressources KM (humaines et techniques) pour les dispositifs d'externalisation et de combinaison des connaissances}{ateliers de retour d'expérience de projets, modélisation des connaissances expertes au sein de communautés de pratique, alimentation des référentiels de connaissances métier, etc. Ceci, pour faciliter l'extraction, la remise en question, la conservation, la validation et la combinaison des connaissances au sein des étapes de vérification et d'ajustement, le «~\pdca{\bf CHECK \&  ACT}{}~» de tous les processus et activités métier ciblés.}
    \end{itemize}
%\end{enumerate}
%\end{enumerate}

\begin{figure}[ht]
    \centering
    \includegraphics[width=\textwidth/2]{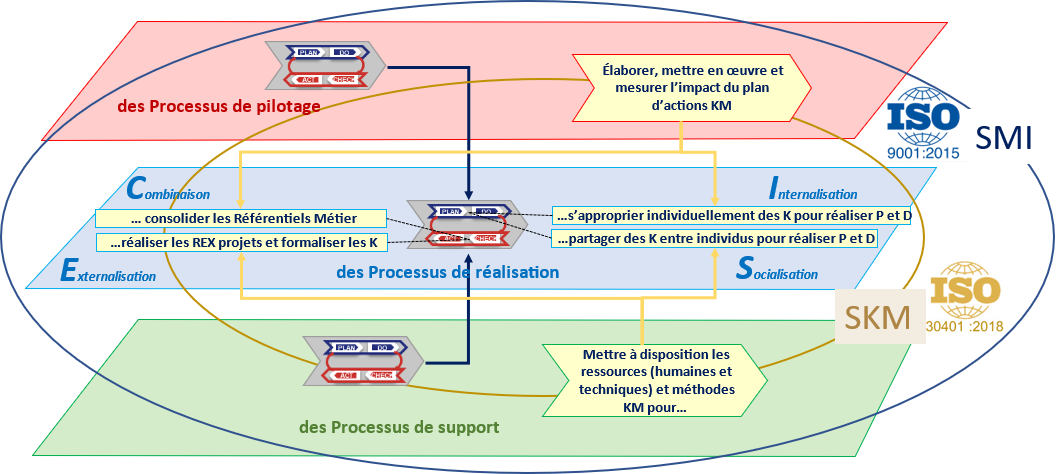}
    \caption{Intégration du Système de Management des Connaissances dans le Système de Management Intégré}
    \label{Fig_ArtApia_11}
\end{figure}

Le schéma \cffig{Fig_ArtApia_11} synthétise le positionnement de ces processus KM de management et de support et leurs interactions avec le déploiement du cycle \lepdca ~dans les processus opérationnels qui conduit à l’intégration des activités et comportements issus du \leseci ~à chaque étape du cycle \lepdca ~:

\begin{itemize}
    \ittguillj{Appropriation}{individuelle par les opérationnels des connaissances mises à disposition par l’organisation ;}
    \ittguillj{Partage} {des connaissances entre opérationnels (pairs de la même discipline ou au contraire praticiens de disciplines complémentaires) ;}
    \ittguillj{Explicitation}{des connaissances collectivement acquises à travers des exercices tels que le REX ou le recueil d’expertise, à la charge des opérationnels à la source de ces connaissances nouvelles ;}
    \ittguillj{Consolidation}{des référentiels d’action des processus et des référentiels métiers par l’intégration de ces connaissances nouvelles (analyse d’impact sur les connaissances explicites existantes, mise à jour des structures de classements, etc.), à la charge et sous la responsabilité des propriétaires de process et responsables métier.}

\end{itemize}

\section{Conclusion}
Suite à l’introduction dans l’\isoQ:2015, de l’exigence de gérer les connaissances organisationnelles, beaucoup d’entreprises, certifiées ou non, se sont posé la question du \sic{comment s’y prendre ?}. Trop souvent avons-nous vu apparaître, dans les cartographies des processus de Systèmes de Management, un simple «~processus KM~», dans la famille des processus Supports, supposé fournir un soutien à la capitalisation et à la diffusion des connaissances. Face à ce constat, cet exposé montre que le KM est bel et bien un système dans lequel on retrouve un processus de pilotage, un processus support et un processus de management des connaissances proprement dit, constitué d’activités intimement intriquées dans les processus opérationnels de l’organisation. Si les processus de pilotage et de support au management des connaissances, aboutissant à la mise en place de gouvernance, méthodes et environnements favorables à l’apprentissage organisationnel, relèvent de professionnels de la discipline, le processus de management des connaissances en lui-même relève des opérationnels métier qui créent, utilisent et mettent à jour ces connaissances. Notre conviction, basée sur les opérations que nous avons conduites depuis 25 ans, est que ce processus peut être mis en place de façon tout à fait naturelle, voire «~indolore~» du point de vue de l’effort à consentir, par l'application du modèle \leseci ~pour la création et l'application des connaissances, dès lors que le cycle \lepdca ~est une pratique effective de l’ensemble des collaborateurs de l’organisation. 

\bibliographystyle{plain}
\bibliography{Biblio_Ardans_SKM_SMI}

\begin{thebibliography}{10}

\bibitem{Argyris2002}
Chris Argyris and Donald~A. Schön.
\newblock {\em Apprentissage organisationnel : théorie, méthode, pratique}.
\newblock De Boeck Université, Paris, 2002.

\bibitem{ArdansIC2023}
Alain Berger.
\newblock Regard sur l'ingénierie de la connaissance face à l'{ISO}30401.
\newblock In {\em 34\up{èmes} Journées francophones d'Ingénierie des Connaissances (IC'2023)}, volume \url{https://lc.cx/Hk8mDx}, Strasbourg, France, July 2023. Plate-Forme Intelligence Artificielle (PFIA).

\bibitem{Crosby1979}
Ph. Crosby.
\newblock {\em Quality is Free: The Art of Making Quality Certain}.
\newblock New York: McGraw-Hill, 1979.

\bibitem{deRosnay75}
Joël de~Rosnay.
\newblock {\em Le Macroscope Vers une vision globale}.
\newblock Éditions du Seuil - Paris, 1975.

\bibitem{Deming}
W.~Edwards Deming.
\newblock {\em Out of the Crisis}.
\newblock MIT Center for Advanced Engineering Study. The MIT Press, 1986.

\bibitem{galinier1989}
Michel Galinier and Pierre Saulent.
\newblock {\em SADT, un langage pour communiquer}.
\newblock Eyrolles, Paris, 1989.

\bibitem{Grundstein2012}
Michel Grundstein.
\newblock Three postulates that change knowledge management paradigm.
\newblock In {\em IntechOpen}, New Research on Knowledge Management Models and Method. Hou H-T, 2012.

\bibitem{IDEF0}
Knowledge Based~Systems Inc.
\newblock IdefØ function modeling method.
\newblock Technical report, KBSI, 1993.

\bibitem{LeMoigne78}
Jean-Louis~Le Moigne.
\newblock {\em La théorie du système général. Théorie de la modélisation}.
\newblock PUF, 1978.

\bibitem{Nishihara-2018CD}
A-H. Nishihara, M.~Matsunaga, I.~Nonaka, and K.~Yokomichi.
\newblock {\em Knowledge creation in community development: Institutional change in Southeast Asia and Japan}.
\newblock Cham, Switzerland: Palgrave Macmillan, 2018.

\bibitem{Nishihara-2018PA}
A-H. Nishihara, M.~Matsunaga, I.~Nonaka, and K.~Yokomichi.
\newblock {\em Knowledge creation in public administrations: Innovative government in Southeast Asia and Japan}.
\newblock Cham, Switzerland: Palgrave Macmillan, 2018.

\bibitem{Nonaka95}
Ikujirō Nonaka and Hirotaka Takeuchi.
\newblock {\em The Knowledge-Creating Company: How Japanese Companies Create the Dynamics of Innovation}.
\newblock Oxford University Press, 1995.

\bibitem{Nonaka19}
Ikujirō Nonaka and Hirotaka Takeuchi.
\newblock {\em The Wise Company How Companies Create Continuous Innovation}.
\newblock Oxford University Press, 1995.

\bibitem{iso2008}
ISO~Central Secretary.
\newblock Iso 9000 introduction and support package: Guidance on the concept and use of the process approach for management systems.
\newblock In {\em ISO/TC 176/SC 2/N 544R3}, International Organization for Standardization. Geneva, CH, 2008.

\bibitem{iso9000}
ISO~Central Secretary.
\newblock Quality management systems — fundamentals and vocabulary.
\newblock In {\em {\url{https://www.iso.org/obp/ui/#iso:std:iso:9000:ed-4:v1:en}}}, International Organization for Standardization. Geneva, CH, 2015.

\bibitem{iso9001}
ISO~Central Secretary.
\newblock Quality management systems — requirements iso9001:2015.
\newblock In {\em {\url{https://lc.cx/nqhkab}}}, International Organization for Standardization. Geneva, CH, 2015.

\bibitem{iumss2018}
ISO~Central Secretary.
\newblock The integrated use of management system standards (iumss).
\newblock In {\em {\url{https://www.iso.org/fr/publication/PUB100435.html}}}. Geneva, CH, 2018.

\bibitem{isoMgt}
ISO~Central Secretary.
\newblock Iso management system standards list.
\newblock In {\em {\url{https://lc.cx/wSL7g_}}}, International Organization for Standardization. Geneva, CH, 2018.

\bibitem{iso30401}
ISO~Central Secretary.
\newblock Knowledge management systems — requirements iso30401:2018.
\newblock In {\em {\url{https://lc.cx/TWDeFT}}}, International Organization for Standardization. Geneva, CH, 2018.

\bibitem{Watzlawick1975}
Paul Watzlawick, John Weakland, and Richard Fisch.
\newblock {\em Changements – Paradoxes et psychothérapie}.
\newblock Paris : Seuil, 1975.

\end{thebibliography}
%\bibliography{Biblio_Ardans}

\end{document}